# Relative Positioning for Aerial Robot Path Planning in GPS Denied Environment.

Corresponding author: Farzad Sanati PhD,
School of Engineering and Technology
CQUniversity Australia
f.sanati@cqu.edu.au

***Abstract***
One of the most useful applications of intelligent aerial robots sometimes called Unmanned Aerial Vehicles (UAV) in Australia is known to be in bushfire monitoring and prediction operations. A swarm of autonomous drones/UAVs programmed to work in real-time observing the fire parameters using their onboard sensors would be valuable in reducing the life-threatening impact of that fire. However autonomous UAVs face serious challenges in their positioning and navigation in critical bushfire conditions such as remoteness and severe weather conditions where GPS signals could also be unreliable.  This paper tackles one of the most important factors in autonomous UAV navigation, namely Initial Positioning sometimes called Localisation. The solution provided by this paper will enable a team of autonomous UAVs to establish a relative position to their base of operation to be able to commence a team search and reconnaissance in a bushfire-affected area and find their way back to their base without the help of GPS signals..

## 6-Introduction

The use of radio frequency (RF) signals for positioning and navigation has advanced significantly in recent years, particularly in scenarios where Global Positioning System (GPS) signals are unavailable or unreliable. Techniques such as Time Difference of Arrival (TDOA) and trilateration have been extensively studied and applied to determine the positions of RF emitters and receivers. This literature review delves into the foundational theories, recent developments, and practical applications of these methods, providing a comprehensive understanding that leads to the proposed solution of calculating the relative location using distances to RF emitters.

## 7-State of the Art

The field of drone path planning and positioning in GPS-denied environments has gained substantial attention in recent years due to the growing applications of unmanned aerial vehicles (UAVs) in various domains. This section reviews significant contributions to the literature, focusing on techniques and methodologies that facilitate effective drone navigation and positioning without relying on GPS.

### 2.1 Doppler-Based Navigation Systems

One of the pioneering works in utilizing Doppler shift for navigation in GPS-denied environments is by Adams et al. (2015). They demonstrated how Doppler radar signals could be used to estimate the velocity and position of UAVs with high accuracy, providing a foundation for subsequent research in Doppler-based navigation systems.

The Doppler effect, which describes the change in frequency or wavelength of a wave in relation to an observer moving relative to the wave source, offers a powerful tool for positioning and navigation. By leveraging the Doppler shift method, it is possible to infer the relative velocity and subsequently the position of a drone based on the observed changes in frequency of signals received from known RF emitters. This technique is particularly valuable in GPS-denied environments, where alternative methods of localization must be employed to ensure operational efficacy (Thrun et al., 2005; Dhillon et al., 2021).

## 2.2 TDOA for Positioning

Time Difference of Arrival (TDOA) is a well-established technique for localizing an RF emitter by measuring the difference in arrival times of a signal at multiple receivers. This method has been widely used in various fields, including telecommunications and navigation.

One of the seminal works in TDOA-based localization is by Chan and Ho (1994), who proposed algorithms for position estimation using TDOA data. Their work laid the groundwork for subsequent research, providing a framework for solving the nonlinear equations involved in TDOA measurements [1].

Recent advancements in TDOA have focused on improving accuracy and computational efficiency. Liu et al. (2020) introduced an improved algorithm that reduces computational complexity while maintaining high accuracy, making it more feasible for real-time applications [2]. Further, Shen et al. (2021) explored the integration of TDOA with other localization techniques, enhancing robustness and precision in multipath environments [3].

In practical applications, TDOA has been employed in systems such as indoor positioning and emergency response. Alarifi et al. (2016) demonstrated the use of TDOA in indoor environments, achieving high precision in tracking mobile devices [4]. Similarly, Xiao et al. (2021) highlighted the use of TDOA in healthcare settings for tracking patients and medical equipment [5].

## 2.3 Trilateration for Localization

Trilateration is another crucial technique for determining the position of a receiver based on known distances to multiple reference points (RF emitters). This method is fundamentally different from TDOA but often used in conjunction.

The basic principles of trilateration are rooted in geometry and have been explored extensively in the context of GPS. Kaplan and Hegarty (2017) provided a detailed explanation of trilateration in their comprehensive book on GPS theory and applications, offering insights into its mathematical foundations [6].

Modern research has focused on enhancing trilateration accuracy in challenging environments. Ning et al. (2017) investigated the use of trilateration in wireless sensor networks, addressing issues such as signal attenuation and noise [7]. Additionally, research by Zhang et al. (2019) on wireless positioning emphasized the importance of integrating trilateration with other localization methods to improve accuracy and reliability in complex scenarios [8].

Practical implementations of trilateration are abundant in the literature. For instance, Han et al. (2018) discussed the application of trilateration in indoor positioning systems, where GPS signals are often obstructed [9]. Moreover, Zhang et al. (2020) explored its use in robotics, where accurate positioning is critical for navigation and task execution [10].

### 2.4 Integration of TDOA and Trilateration

Integrating TDOA and trilateration can significantly enhance the reliability and accuracy of positioning systems, especially in environments with limited GPS availability.

Combining TDOA with trilateration has been proposed to leverage the strengths of both methods. He et al. (2020) developed a hybrid algorithm that integrates TDOA and trilateration for improved localization accuracy in wireless networks [11]. Their approach demonstrated significant improvements in scenarios with limited infrastructure.

In practical applications, such integrated systems have shown promising results. For example, a study by Wang et al. (2020) demonstrated the use of combined TDOA and trilateration in vehicular networks, achieving robust localization despite signal interference [12]. Similarly, Li et al. (2021) applied these techniques to drone navigation in GPS-denied environments, providing a reliable solution for positioning and tracking [13].

### 2.5 Challenges and Future Directions

While TDOA and trilateration are powerful techniques, they face several challenges, including multipath propagation, signal interference, and non-line-of-sight conditions. Addressing these issues is crucial for advancing the state of the art in RF-based positioning.

In summary, the literature on TDOA and trilateration provides a comprehensive framework for understanding and implementing RF-based positioning systems. By integrating these techniques and addressing the associated challenges, it is possible to develop robust solutions for determining relative locations in GPS-denied environments. Continued research and technological advancements will further enhance the accuracy and reliability of these systems, paving the way for their widespread adoption in various applications.

## 8-Discussion

**Theorem 1:**

There is a RF receiver that is receiving RF from an Emitter from an unknown location, also receiving the reflected RF from a moving reflector we can calculate the distance between the receiver and the RF emitter using Doppler shift.

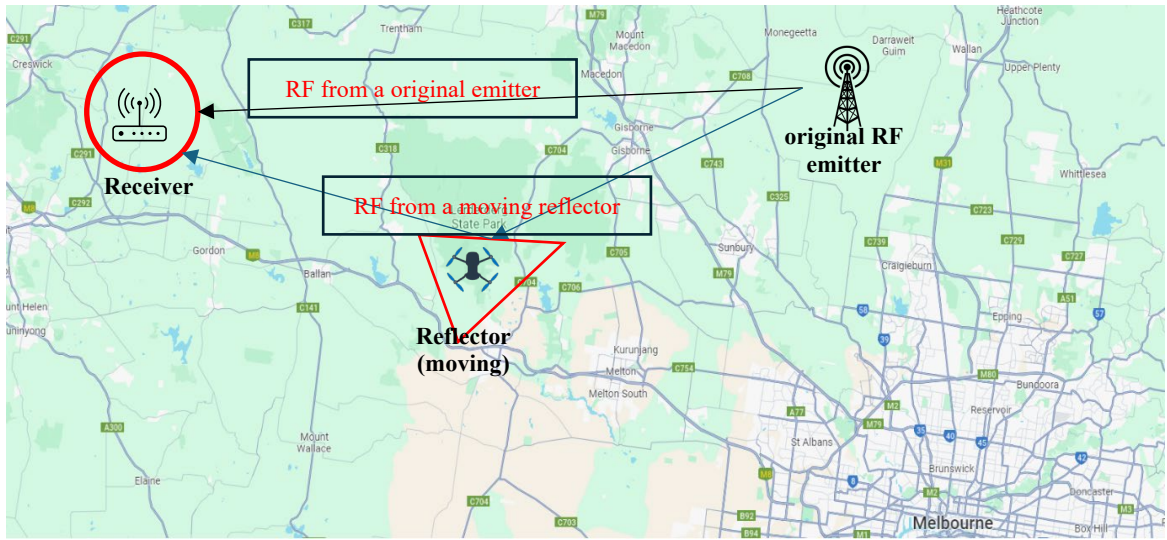

Figure 1: Depicting passive radar using Doppler Shift, (source Google Maps)

If the emitter is moving relative to the receiver, here's how the ideal scenario would work (reflected signal not considered):

### Doppler Shift:

The formula for Doppler shift ($f_d$) is: $f_d = |f_{received} - f_{emitted}|$

where:

- $f_d$ = Doppler shift (Hz)
- $f_{received}$ = Frequency of the received signal (Hz)
- $f_{emitted}$ = Frequency of the emitted signal (Hz)

### Distance Estimation (assuming negligible relative velocity):

In a stationary scenario, we can estimate the distance (d) using the simplified formula:

$$d = \frac{(c \cdot f_d)}{(2 \cdot f_{emitted})}$$

where c = Speed of light (approximately 3 x 10^8 meters per second)

### Challenges with Reflected Signal:

The presence of a reflected signal significantly complicates the scenario. The reflected signal will also experience a Doppler shift due to its movement relative to the receiver, creating an overlapping effect. Separating the Doppler shift caused by the emitter's motion from the reflected signal's Doppler shift is nearly impossible.

Therefore, using Doppler shift alone isn't sufficient for accurate distance measurement in this case.

**Alternative Approaches:**

- **Time Difference of Arrival (TDOA):** This technique measures the time difference between the direct and reflected signals to estimate the position of the emitter.
- **Multiple receivers:** By using multiple receivers at known locations, triangulation techniques can be employed to pinpoint the emitter's position based on the received signal's arrival times.

## 2.6 Using Multiple Receivers

**Theorem 2:**

Multiple receivers can be used to calculate the distance between the emitter and a receiver using Time Difference of Arrival (TDOA):

To calculate the general direction of an RF emitter from the perspectives of three receivers, you can use a technique called Time Difference of Arrival (TDOA). Here's a step-by-step outline of the process:

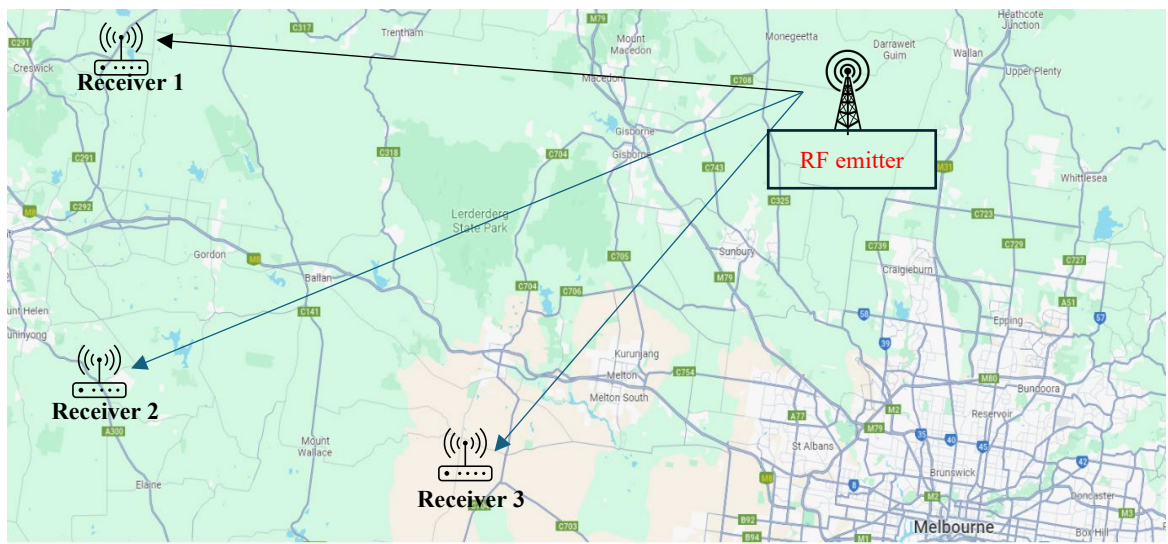

Figure 2: Depicting multi-receiver. (source Google Maps)

1. Setup and Assumptions:
    - Place three receivers (let's call them $R_1, R_2, and\ R_3$ at known positions $(x_1, y_1), (x_2, y_2), and\ (x_3, y_3)$.

    - The RF emitter is located at an unknown position *(x, y)*.

2. Time Difference of Arrival (TDOA):
    - Each receiver detects the signal emitted by the RF source.
    - Record the arrival times of the signal at each receiver: $t_1, t_2, and\ t_3$.
    - Calculate the differences in arrival times:
        $\Delta t_{1,2} = t_1 - t_2$
        $\Delta t_{1,3} = t_1 - t_3$

3. Convert Time Differences to Distance Differences:
    - Assuming the signal travels at the speed of light $c$ :
    $$D_{1,2} = c \cdot \Delta t_{1,2}$$

    - $d_{1,2}$ and $d_{1,3}$ are the differences in distances from the emitter to the receivers.

4. Hyperbolic Positioning:
    - The differences in distances define hyperbolic curves on which the emitter must lie.
    - For $R_1$ and $R_2$ :

    $$\sqrt{(X - x_1)^2 + (y - y_1)^2} - \sqrt{(x - x_2)^2 + (y - y_2)^2} = d_{1,2}$$

    - For $R_1$ and $R_3$ :

    $$\sqrt{(X - x_1)^2 + (y - y_1)^2} - \sqrt{(x - x_3)^2 + (y - y_3)^2} = d_{1,3}$$

5. Solving the System of Equations:
    - Solve these two nonlinear equations simultaneously to find the coordinates *(x, y)* of the RF emitter. This typically requires numerical methods or iterative algorithms due to the complexity of the equations.

6. General Direction Calculation:
    - Once the position *(x, y)* of the RF emitter is determined, calculate the direction vector from each receiver to the emitter:
    $$\vec{D}_i = (x - x_i, y - y_i)$$

    - Normalize the direction vectors if needed:
    $$\hat{D}_i = \frac{\vec{D}_i}{|\vec{D}_i|}$$

7. Combining Directions:
    - If a general direction is needed (e.g., for navigation purposes), combine the direction vectors from the three receivers. One method is to average the unit vectors:

    $$\hat{D}_{avg} = \frac{\hat{D}_1 + \hat{D}_2 + \hat{D}_3}{3}$$

    - The resulting vector $\hat{D}_{avg}$ gives a general direction towards the RF emitter.

This method provides a way to triangulate the position and direction of the RF emitter using the time differences of signal arrivals at three known points.

## 2.7 Using Multiple RF Emitter with Multiple Receivers

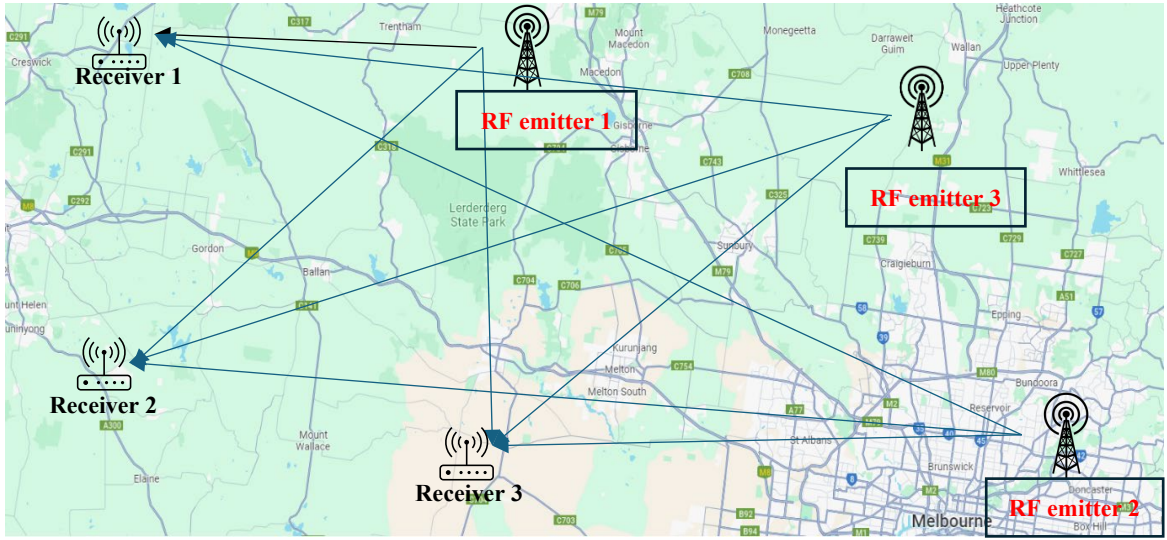

Figure 3: Depicting multi-receiver and multi-emitter. (source Google Maps)

**Theorem 3:**
We can calculate the relative location by determining the distances to three known RF emitters using a method called "trilateration". Here's how you can do it:

**The Process**
1. Known Positions of RF Emitters:
    - Let the positions of the three RF emitters be $E_1(x_1, y_1), E_2(x_2, y_2), and\ E_3(x_3, y_3)$.

2. Measured Distances to Emitters:
    - Measure the distances from your location to each of the RF emitters:
        $d_1, d_2, and\ d_3$.

3. Formulate Distance Equations:
    - For each emitter, you can write an equation based on the Euclidean distance formula:

$$(x - x_1)^2 + (y - y_1)^2 = d_1^2$$

$$(x - x_2)^2 + (y - y_2)^2 = d_2^2$$

$$(x - x_3)^2 + (y - y_3)^2 = d_3^2$$

4. Solve the System of Equations:
    - You need to solve this system of nonlinear equations to find the coordinates *(x, y)* of your location. Here's a simplified example:

**Example Calculation**
Assume you have three RF emitters with known positions and measured distances as follows:
  $E_1(0,0) with\ d_1 = 5$
  $E_2(10,0) with\ d_2 = 5$
  $E_3(5,10) with\ d_3 = 5$

1. Equations:
    - From $E_1$:

$$x_2 + y_2 = 25$$
- From $E_2$:
$$(x - 10)^2 + y_2 = 25$$
- From $E_3$:
$$(x - 5)^2 + (y - 10)^2 = 25$$

2. Simplify and Solve:
   - Substitute the equations into each other and solve step-by-step:
     - From $E1$:
$$x_2 + y_2 = 25$$
     - From $E2$:
$$(x - 10)^2 + y_2 = 25 \rightarrow x_2 - 20x + 100 + y_2 = 25 \rightarrow x_2 + y_2 - 20x + 100 = 25 \rightarrow x_2 + y_2 = 25 \rightarrow 20x + 100 = 0 \rightarrow x = 5$$
     - From $E3$:
$$(5 - 5)^2 + (y - 10)^2 = 25 \rightarrow y_2 - 20y + 100 = 25 \rightarrow y_2 - 20y + 75 = 0$$
+
   - Solve the quadratic equation for $y$:
$$y = 15 \text{ or } y = 5$$

3. **Result:**
   - Given the distances, the solutions would be:
     *(x, y) = (5, 15) or (5, 5)*

   - Verify with all three equations to determine the correct position. For this example, *(x, y) = (5, 5)* satisfies all the equations.

**General Approach**

**Nonlinear Systems:** For more complex and realistic scenarios, solving the nonlinear system might require numerical methods or iterative algorithms like the Newton-Raphson method.
**Multidimensional Trilateration:** The above method can be extended to three dimensions by adding the z-coordinate and using three-dimensional distance equations. Using this approach, you can determine your relative location with respect to three known RF emitters based on the measured distances to them.

# 6-Multidimensional Trilateration

Our two-dimensional multi-receiver method can be extended to three dimensions by adding the z-coordinate and using three- dimensional distance equations. We can use this approach to determine our relative location concerning three known RF emitters based on their measured distances. So far, we discussed the multi-receiver method that can calculate the relative average distance of receivers from three RF emitters. To achieve more accuracy, we consider three drones each at a different altitude scattered apart from each other, each drone has an SDR, and all three SDRs are using the same clock on the network. We can calculate the relative location using three-dimensional Trilateration by determining the distances to three known RF emitters.
**Theorem 4:**
**Step1 -** Finding Emitter Positions Using TDOA Given: - Three drones with known positions where $i \in \{1,2,3\}$- Arrival times of signals at these drones $t_i$ where $i \in 1,2,3$ - Speed of light $c$.

Time Differences of Arrival

$$\Delta t_{12} = t_1 - t_2$$
$$\Delta t_{13} = t_1 - t_3$$

Convert Time Differences to Distance Differences
$$d_{12} = c \cdot \Delta t_{12}$$
$$d_{13} = c \cdot \Delta t_{13}$$

Hyperbolic Positioning Equations

$$\sqrt{(X-x_1)^2 + (y-y_1)^2 + (z-z_1)^2} - \sqrt{(x-x_2)^2 + (y-y_2)^2 + (z-z_2)^2} = d_{1,2}$$

- For $R_1$ and $R_3$:

$$\sqrt{(X-x_1)^2 + (y-y_1)^2 + (z-z_1)^2} - \sqrt{(x-x_3)^2 + (y-y_3)^2 + (z-z_3)^2} = d_{1,3}$$

We minimize the squared differences to find the emitter position:

$$f(x,y,z) = (\sqrt{(x-x_1)^2 + (y-y_1)^2 + (z-z_1)^2} - \sqrt{(x-x_3)^2 + (y-y_3)^2 + (z-z_3)^2} - d_{12})^2 + (\sqrt{(x-x_1)^2 + (y-y_1)^2 + (z-z_1)^2} - \sqrt{(x-x_3)^2 + (y-y_3)^2 + (z-z_3)^2} - d_{13})^2$$

**Step2 - Finding Relative Average Location Using Trilateration** Given: - Estimated positions of the emitters ($x_{ei}, y_{ei}, z_{ei}$) where $i \in \{1, 2, 3\}$. - Known positions of the drones ($x_{di}, y_{di}, z_{di}$) where $i \in \{1, 2, 3\}$. Distances

$$d_{i,j} = \sqrt{\left(x_{d_i} - x_{e_j}\right)^2 + \left(y_{d_i} - y_{e_j}\right)^2 + \left(z_{d_i} - z_{e_j}\right)^2}$$

Trilateration to Find Relative Position We solve for the coordinates $(x, y, z)$ that minimize:

$$f(x,y,z) = \sum_{i=1}^{3} \left( \sqrt{\left(x - x_{e_i}\right)^2 + \left(y - y_{e_i}\right)^2 + \left(z - z_{e_i}\right)^2} - d_i \right)^2$$

In three-dimensional space, trilateration involves determining the position of a point relative to three known positions and their respective distances. Given three emitters with known coordinates ($E_1, E_2, E_3$) and their distances ($d_1, d_2, d_3$) to the point $(x, y, z)$, the system of equations is:

$$(x - x_1)^2 + (y - y_1)^2 + (z - z_1)^2 = d_1^2$$
$$(x - x_2)^2 + (y - y_2)^2 + (z - z_2)^2 = d_2^2$$
$$(x - x_3)^2 + (y - y_3)^2 + (z - z_3)^2 = d_3^2$$

# 7-Experiment and Simulation

This experiment aims to determine the average relative location of three Software Defined Radio (SDR) receivers by calculating the distances to three RF emitters using trilateration. This experiment will allow us to accurately determine the average relative location of the three receivers based on their distances to the three unknown RF emitters. By visualizing the results, we can validate the effectiveness of the trilateration method and refine the setup or calculations if needed.

Let's assume the positions of the emitters are as follows:
$$E_1 = (0,0,0), E_2 = (500,0,0), E_3 = (0,500,0)$$
And the distances from the point to each emitter are:
$$d_1 = 300, d_2 = 400, d_3 = 500$$
We will solve the system of equations using these values.
$$x^2 + y^2 + z^2 = 300^2$$
$$(x - 500)^2 + y^2 + z^2 = 400^2$$
$$x^2 + (y - 500)^2 + z^2 = 500^2$$

From the second equation:

$$x^2 + y^2 + z^2 - 1000x + 250000 = 160000$$
$$x^2 + y^2 + z^2 - 1000x = -90000$$
We already know $x^2 + y^2 + z^2 = 90000$ from the first equation:
$$x^2 + y^2 + z^2 - 1000y + 250000 = 250000$$
$$x^2 + y^2 + z^2 - 1000y = 0$$

Substitute (x = 180) and (y = 90) back into the first equation to find (z):
$$180^2 + 90^2 + z^2 = 90000$$
$$z^2 = 49500$$
$$z = \sqrt{49500} \approx 222.36$$

**Results**

The coordinates of the point are approximately:
$$(x, y, z) = (180, 90, 222.36)$$
Figure 4 is showing the simulation results of the data set implemented in Python code.

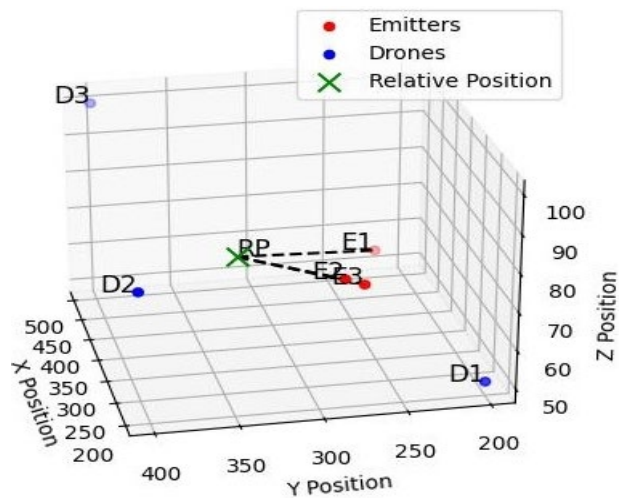

Figure 4: Simulation results using python.

# 8- Conclusion

Relative positions and directions using RF signals is a powerful technique in the realm of navigation and positioning, particularly in GPS-denied environments. By employing Time Difference of Arrival (TDOA) methods, we can determine the position and direction of an RF emitter relative to multiple receivers. This involves solving a system of nonlinear equations based on the differences in signal arrival times at known receiver locations.

Moreover, trilateration can be used to determine a receiver's position by measuring distances to three or more known RF emitters. This method, which relies on solving Euclidean distance equations, is fundamental in various applications, including indoor positioning systems and autonomous navigation.

Combining TDOA and trilateration will ensure a to acquire sufficient information for establishing a relative base position for a team of aerial robots. This method is sufficient to accurately measure distance of a team of aerial robots from three RF emitters, hence being able to pinpoint the relative position of the aerial robot team. These techniques, while mathematically intensive, provide robust solutions for determining relative positions in GPS-denied environment therefore provide sufficient information for path planning and directions in complex environments. As technology advances, these methods will continue to play a crucial role in enhancing the precision and reliability of location-based aerial robots.

Future research should focus on developing algorithms that can mitigate the effects of multipath and non-line-of-sight conditions. Wang et al. (2019) on wireless communications provide a foundation for understanding these challenges and suggest potential solutions [14]. Additionally, advancements in machine learning and artificial intelligence offer new avenues for enhancing the accuracy and robustness of localization systems [15].

Emerging technologies, such as ultra-wideband (UWB) and 5G, hold promise for improving TDOA and trilateration accuracy. Decarli et al. (2021) explored the use of UWB for high-precision localization, highlighting its potential in both indoor and outdoor environments [16]. Moreover, 5G's high bandwidth and low latency characteristics make it an attractive candidate for future positioning systems [17].

## 9-References